\documentclass[11pt]{article}
\usepackage[a4paper,margin=2.3cm]{geometry}
\usepackage{libertinus}
\usepackage{amsmath}
\usepackage[authoryear,round]{natbib}
\usepackage{graphicx,longtable,booktabs,array,calc,etoolbox}
\usepackage[hidelinks]{hyperref}
\usepackage{xcolor}
\providecommand{\tightlist}{\setlength{\itemsep}{0pt}\setlength{\parskip}{0pt}}
\AtBeginEnvironment{longtable}{\small}
\makeatletter\def\verbatim@font{\footnotesize\ttfamily}\makeatother
\title{RL Can Only Reinforce What It Sees Succeed}
\author{Ruben Sanchez-Corcuera \quad Unai Zulaika \quad Aritz Bilbao-Jayo \quad Miguel Fernandez-de-Retana\\[3pt]University of Deusto\\[2pt]\texttt{\{ruben.sanchez, unai.zulaika, aritz.bilbao, m.fernandezderetana\}@deusto.es}}
\date{}
\begin{document}
\maketitle
\begin{abstract}
When a familiar rule stops working, an agent has to weigh what pre-training taught it against what it now observes. We ask how outcome-reward reinforcement learning (RL) changes this balance, in a puzzle world where both sides are known: small language models trained from scratch on games played by a Bayesian player, so that the prior and the right answer are exact. Base models hedge as their corpus did, and RL turns that hedging into decisions, in whichever direction the reward favours. But RL can only reinforce what it sees succeed. When the right action is rare, each run either learns it everywhere or not at all, and which runs take off is predicted before training by one count: the expected number of groups of attempts in which the action wins. About 35 such groups give even odds. The count holds for new and repeated problems and for groups of 4 to 256 attempts, and across learning rates, optimizers, GRPO variants and model sizes from 1.3M to 85M parameters, which move the threshold but not what is counted.
\end{abstract}
\subsection{1. Introduction}\label{introduction}

Imagine an agent that has learned, from thousands of games, that a key opens a locked chest. It turns the key and nothing happens. Should it try again? Keys sometimes jam or go in the wrong way, so a single failure says little. After three failures in a row, though, it becomes more likely that something has really changed, that the key or the lock is broken, and that some other tool now does the job. A good agent weighs what it learned before, its prior, against what it has just seen, the evidence, and switches when the evidence wins.

Language-model agents face this conflict all the time. A coding agent calls a library function the way it saw it in pre-training, and the call fails: was it a typo, a flaky test, or a new version of the library that changed the interface \citep{liu2024codeupdatearena}? A web agent clicks where a button used to be and nothing happens: the page may still be loading, or it may have been redesigned. An assistant with long-term memory acts on what it stored about a user, but the user has moved or changed their mind, and the memory itself is now wrong \citep{chen2026compliance,hu2026memoryagentbench}. In each case a single failure is ambiguous, and the agent must decide how much evidence it takes to stop trusting what it knows. More and more of these agents are shaped by reinforcement learning (RL) with outcome rewards \citep{lambert2024tulu3,guo2025deepseekr1}. So we ask a simple question: how does this post-training change who wins? Two answers sound equally plausible. RL could make the agent more stubborn, because the familiar action is the one that usually earns the reward. Or it could make the agent more responsive, because switching at the right moment also earns the reward.

The question is hard to answer with pre-trained language models alone. We do not know how strong their prior is, and when the evidence is noisy we do not know what the right answer would be. Most studies of knowledge conflicts give a model one question and one contradicting document \citep{longpre2021entity,xu2024survey}. Recent work lets the model act \citep{xu2026code,chen2026compliance}, but it neither controls the strength of the prior nor compares the same model before and after RL.

We therefore build a world in which both sides of the conflict are known. Puzzle rules are data, and we train small GPT models from scratch on 9 million games played by a Bayesian player. This lets us set exactly how often each rule is seen and how often it changes, and it tells us, at every point of a game, what an ideal agent would do. We then post-train the models with outcome-reward RL and measure exactly how their willingness to switch changes. The story that emerges has three steps.

\textbf{RL turns hedging into decisions.} Before RL, a model hedges the way its training data did: after each failure it switches about as often as the Bayesian player would. RL replaces this hedging with decisions. The agent switches when switching is the better bet and holds on when it is not, and it moves toward the optimal policy in both directions.

\textbf{But only if the right action is already tried.} RL can only reward an action that the agent takes. When we make switching rare in the training corpus, RL does not discover it, and it makes switching rarer still. Near this boundary, each RL run is all or nothing: it either learns to switch for every rule at once, or it learns nothing.

\textbf{Whether a run takes off can be predicted by counting.} RL learns from groups of attempts at the same problem, and a group can teach the new action only if one of its attempts uses it and wins. Before training, the base model tells us how many such groups a run can expect. That one number predicts which runs take off: about 35 expected successful groups give even odds (Figure 1). It does not matter whether the groups come from new problems or repeated ones, or how many attempts each group holds. The learning rate, the optimizer and the size of the model change how many groups are needed, but not what has to be counted. The count also changes what ``rare'' means: a switch that RL never learned with our budget was learned with four to eight times more, as often as the count predicted.

Who wins, then, is not decided by RL itself. Whether post-training makes an agent stubborn depends on how often the right action succeeds under the base model, and on how much experience RL gets to see it succeed.

In summary, we make four contributions:

\begin{itemize}
\tightlist
\item
  \textbf{A world where the prior is known.} A puzzle environment and small language models trained from scratch on games played by a Bayesian player, in which the strength of the prior, the evidence and the optimal decision are all exact, together with an exact measure of how a model weighs them before and after RL.
\item
  \textbf{RL makes agents decisive, not stubborn.} Base models hedge like their corpus, and outcome-reward RL moves their decisions toward the optimal policy in both directions: it switches more where switching pays and less where it does not.
\item
  \textbf{When the right action is rare, each run is all or nothing.} RL sharpens the prior when the switch is rare, and near this boundary a run either learns the switch for every rule or for none.
\item
  \textbf{A count that predicts take-off before training.} The expected number of groups in which the new action wins, computed from the base model, predicts which of 400 runs take off, for new and repeated problems and for groups of 4 to 256 attempts. Across twelve changes of recipe (910 runs), the recipe moves the threshold but not what is counted.
\end{itemize}

We will release our code and data as open source.

\begin{figure}[tbp]
\centering
\includegraphics[width=0.62\textwidth]{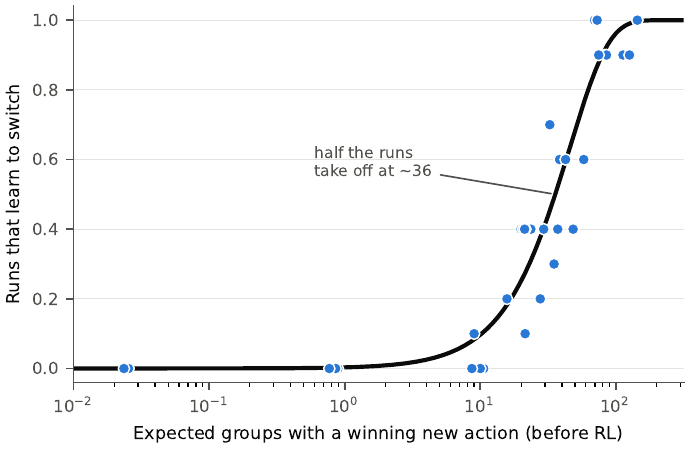}
\caption{\textbf{Counting successful groups predicts whether RL learns the new action.} Each point is a set of 10 RL runs that start from a base model (400 runs). The horizontal axis is the number of groups in which at least one attempt would take the new action and win, expected over the whole run and computed from the base model before training. The vertical axis is the share of runs that learn to switch. The line is the law fitted to all 400 runs.}
\label{fig:teaser}
\end{figure}

\subsection{2. A world where we know the prior}\label{a-world-where-we-know-the-prior}

\textbf{A game with an ambiguous failure.} Our agent plays short text puzzles. It stands in a room whose exit is blocked by one to three obstacles, such as a locked chest. Next to each obstacle lie two candidate tools, for example a key and a hammer. The agent can take a tool, use a tool on an obstacle, and leave once every obstacle is gone; the game lists the valid actions at each step and pays a reward of 1 if the agent leaves within the step limit. There are 9 rules we test and 30 filler rules, and each one pairs an obstacle with its default tool (\emph{D}): the key opens the chest, the torch melts the ice wall.

In the normal world the default tool works 80 \% of the time, and the other candidate does nothing. In a changed world, as when a lock breaks in the real world, the default never works, and the other candidate, the alternative (\emph{A}), removes the obstacle. Every failure produces the same observation, ``nothing'', so a failed key is ambiguous: it may have jammed, or the lock may have changed. The alternative is always the default tool of some other rule, so the hammer is a perfectly ordinary tool; only the learned link between chest and key tells the two apart (Figure 2a).

\begin{figure}[tbp]
\centering
\includegraphics[width=\textwidth]{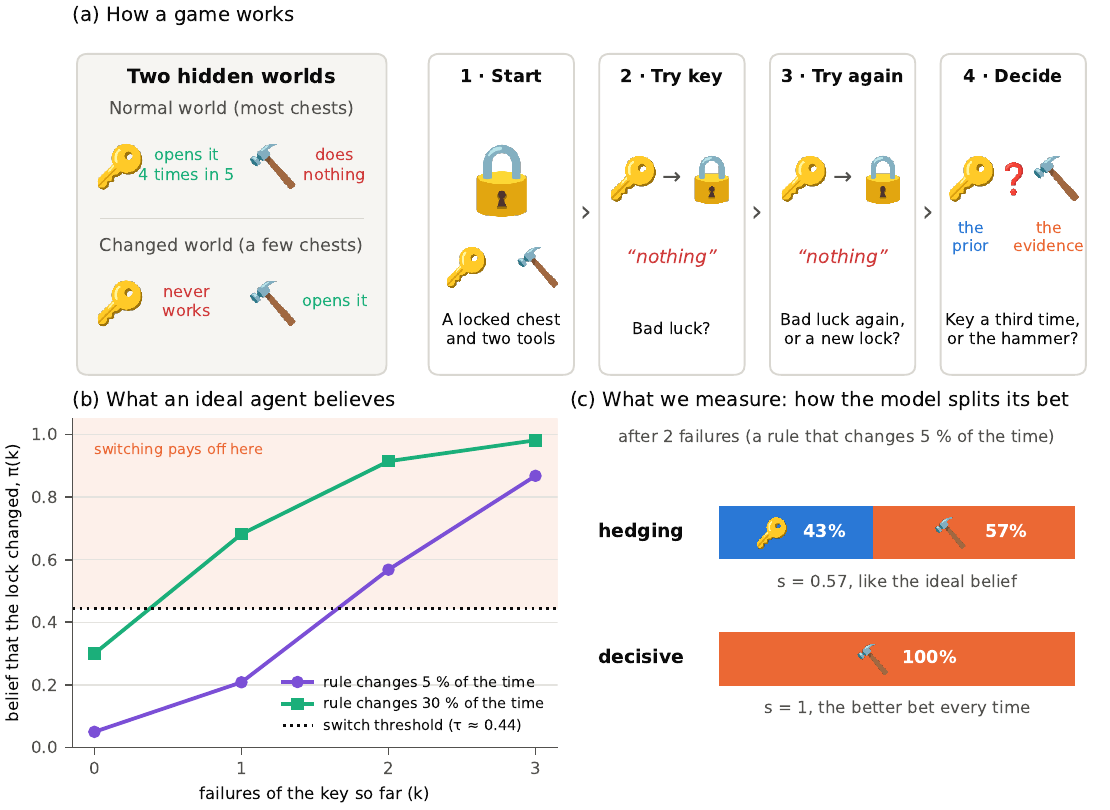}
\caption{\textbf{The game, what an ideal agent believes, and what we measure.} (a) Each chest follows one of two hidden worlds. The agent tries the key, which usually works; after a few failures it must choose between trying again and switching to the hammer. (b) The ideal agent's belief that the lock has changed grows with each failure, faster for a rule that changes more often; above the dotted line, switching is the better bet. (c) We measure how a model splits its probability between the two tools. A model that hedges like its corpus puts 57\,\% on the hammer after two failures; a decisive one puts everything on the better bet. Icons: Noto Emoji.}
\label{fig:world}
\end{figure}

\textbf{What an ideal agent would do.} Suppose each obstacle follows its changed rule with probability \(\nu\), the rule's unreliability. An ideal agent starts by believing the rule changed with probability \(\nu\) and updates after each failure. After \emph{k} failures of the default tool, its belief is the Bayesian posterior

\[\pi(k) = \frac{\nu}{\nu + (1 - \nu)\, 0.2^{k}},\]

where 0.2 is the chance that the default fails by bad luck. With \(\nu = 0.05\), one failure raises the belief from 5 \% to 21 \%, a second to 57 \%, a third to 87 \% (Figure 2b). Switching is the better bet once \(\pi(k)\) passes \(\tau \approx 0.44\), so here the ideal agent tries the key twice and then reaches for the hammer.

For a model, we measure the same quantity from the other side. At a decision point like Figure 2a, we take the probability the model gives to each tool and compute its \emph{switch probability}

\[s(k) = \frac{p(A)}{p(A) + p(D)}.\]

A model that hedges like the ideal believer has \(s(k) \approx \pi(k)\). A model that decides has \(s(k)\) near 0 below the threshold and near 1 above it (Figure 2c). Everything we report is a comparison of \(s(k)\) with \(\pi(k)\).

\textbf{Models that learn only from games we wrote.} To know the prior, we write the whole pre-training corpus ourselves. Each corpus holds 9 million games, rendered in a small templated language with one token per word, and played by a scripted Bayesian player: it tries the default first and, after \emph{k} failures, switches with probability \(\pi(k)\). A model that learned this corpus perfectly would hedge exactly like the ideal believer. On each corpus we train a small GPT from scratch \citep{radford2019language,karpathy2022nanogpt}, with 6 layers and 10.7M parameters, for a single pass, so that every game is seen exactly once.

Because we write the corpus, we can turn three knobs that are hidden in a real pre-trained model. \emph{Exposure} sets how often a rule appears. \emph{Reliability} sets how often the rule changes, through \(\nu\). And \emph{rarity} sets how often the corpus ever shows the right switch. For this last knob we mix two players. A share \(\rho\) of the games is played by the Bayesian player; the rest by a late player who keeps trying the default until it has failed four times, too late to matter. The player is drawn per game and plays every obstacle in it, filler rules included, so the general habit of switching early is as rare in the corpus as the switch on the test rules. The smaller \(\rho\), the less often the corpus shows the hammer at the moment it should be used: from once in 23 decisions after two failures down to once in 23,000. Every measurement is exact: we score each valid action by its probability under the model instead of sampling, and we compare each model after RL with its own base model on the same decision points.

\textbf{Outcome-reward RL, and why groups matter.} We post-train with GRPO \citep{shao2024deepseekmath}, without a KL penalty and without dividing the advantage by the group's spread \citep{liu2025understanding}. Each problem is one obstacle in a world drawn from the same distribution as pre-training. The model plays \emph{N} attempts at the problem; each attempt's advantage is its reward minus the mean reward of its group, and one gradient step follows each batch. A step limit of four to seven moves makes the decision after a few failures a last chance, where the reward-maximizing choice is exactly to switch when \(\pi(k)\) passes \(\tau\). Unless stated otherwise, a run takes 200 steps of 512 attempts.

One property of this algorithm drives the rest of the paper. If every attempt in a group fails, or every attempt succeeds, all advantages are zero and the group teaches nothing. A group pushes the model toward the hammer only when at least one attempt actually switches and wins.

\textbf{When a run takes off.} We say that RL has taught a rule when the model's switch probability for that rule, after two or after three failures, rises by more than 0.05 above the base model, beyond the noise of the measurement. A run \emph{takes off} when it learns more than half of the nine rules. This is rarely a close call: as Section 3 shows, a run almost always learns all nine rules or none.

\subsection{3. RL turns hedging into decisions, if it sees the switch}\label{rl-turns-hedging-into-decisions-if-it-sees-the-switch}

\textbf{Base models hedge like their corpus.} We start with the models before any RL. When a base model has seen the key fail twice, how does it split its bet between the key and the hammer? Almost exactly as the ideal believer would (Figure 3a). For a rule that changes 5 \% of the time, the model puts 22 \% on the hammer after one failure and 56 \% after two; the Bayesian posterior says 21 \% and 57 \%. For a rule that changes 30 \% of the time, the whole curve sits higher, again on top of the posterior. Across all conditions, a well-exposed model is off by less than two percentage points on average.

The model learns more than a single habit per rule. It tracks how reliable each rule is: rules that change more often get a higher switch probability after the same evidence, and with high exposure the model reproduces 95 \% of the gap that the posterior predicts between a reliable and an unreliable rule. It also tracks how much evidence it has seen, since its switch probability climbs with every failure. How often a rule was seen matters less, and in a surprising direction: rules seen in only 900 games produce a \emph{flatter} curve, one that moves less with each failure. A model that has seen a rule rarely hedges more, not less.

\begin{figure}[tbp]
\centering
\includegraphics[width=\textwidth]{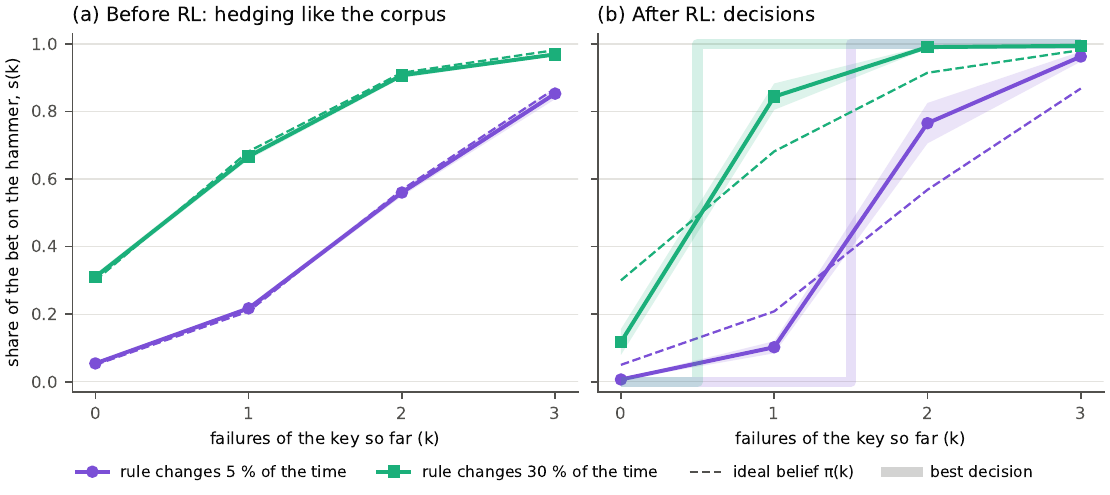}
\caption{\textbf{RL turns hedging into decisions.} Share of the bet on the hammer after $k$ failures of the key, for a rule that changes 5\,\% (purple) or 30\,\% (green) of the time; mean of 18 models, with a 95\,\% band. Dashed: the ideal belief. (a) Base models follow the ideal belief almost exactly. (b) After outcome-reward RL, the same models move toward the best decision (thick pale steps): they stop switching where switching does not pay, and switch more where it does. They now depart from the ideal belief, in the direction of the reward.}
\label{fig:decisions}
\end{figure}

\textbf{RL makes the agent decisive, not stubborn.} What should outcome-reward RL do to such a model? One might expect it to strengthen the prior. Most of the time the key works, so most rewarded games are games in which the agent used the key; RL should push the agent toward the familiar action and make it slower to switch. It does the opposite. In all six combinations of objective and group size we tried, the decisions where prior and evidence conflict moved toward the better choice, not toward the familiar one. Figure 3b shows the pattern. After one failure of a reliable rule, the base model put 22 \% on the hammer; after RL it puts 10 \%, because one failure is not yet enough reason to switch. After two failures, when switching has become the better bet, it puts 77 \% instead of 56 \%. For an unreliable rule the change runs the other way: after a single failure, the share on the hammer rises from 67 \% to 84 \%. Every condition moves toward the best decision, whichever side that is.

RL does not make the model a better Bayesian. Its bets now depart from the ideal belief, in both directions, because the reward does not pay for calibrated beliefs; it pays for good choices. And the choices are good: on held-out problems, RL closes 74 \% to 95 \% of the gap in reward between the base model and the optimal policy. Rules that were rarely seen in pre-training catch up as well.

\textbf{When the switch is rare, RL sharpens the prior.} In all of these models the hammer was never a strange choice: the corpus showed it often enough that the base model already tried it at the right moments. We now turn the rarity knob. We keep a reliable, well-exposed rule, where switching after two failures is the better bet, and lower how often the corpus shows the switch. The base model learns what it sees: after two failures, it reaches for the hammer about as often as the corpus did. We then run RL exactly as before, with 10 base models per level of rarity and groups of 4, 16 or 64 attempts on the same budget.

When the hammer is shown often enough, RL behaves as above: with the switch in 1 of 23 corpus decisions, 9 of 10 runs with small groups learn to switch, and RL closes 83 \% of the gap to the optimal policy. When the switch appears in 1 of 232 decisions or less, no run learns it, out of 90. RL does worse than nothing here. It cannot reward an action it does not take, but it keeps rewarding the key, which usually works, and so the rare switch becomes rarer still: RL removes between a quarter and a half of the little probability the base model gave the hammer.

\begin{figure}[tbp]
\centering
\includegraphics[width=\textwidth]{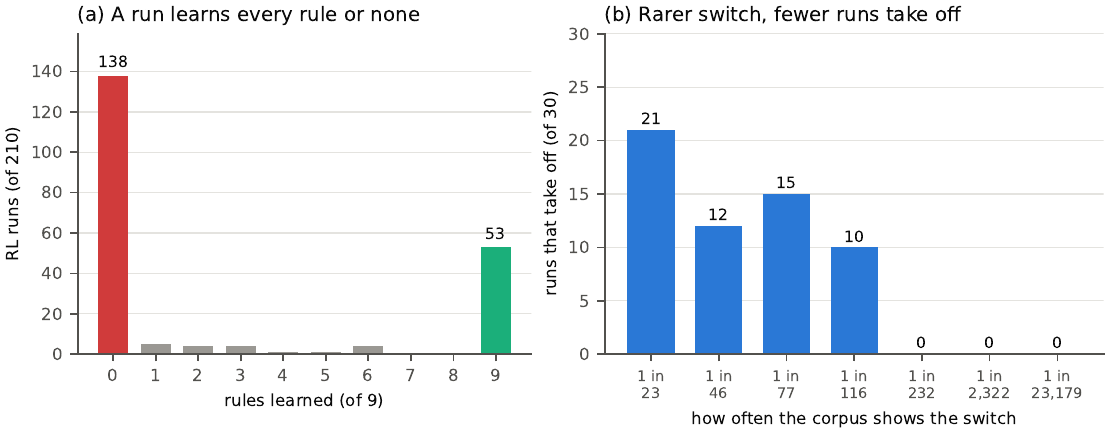}
\caption{\textbf{When the switch is rare, each run learns it everywhere or not at all.} (a) How many of the nine rules each of 210 RL runs learned to switch on. (b) Runs that take off, of 30 (ten base models and three group sizes), against how often the corpus shows the switch after two failures.}
\label{fig:rare}
\end{figure}

\textbf{Each run is all or nothing.} Between these two extremes, the unit that matters is the run, not the decision. Of 210 runs, 138 learn no rule, 53 learn all nine, and only 19 fall in between (Figure 4a). When a run takes off, it learns a general rule, ``after the default fails, try the other tool'', and applies it to all nine rules at once, including decisions where it almost never tried the hammer. In one run, RL never tried the hammer on one of the rules, and yet that rule's switch probability rose by 93 percentage points. This does not contradict the title of this paper. What RL reinforces here is not one tool on one chest but a skill, ``switch after the default fails'', and that skill has to be seen to succeed often enough across the run's problems. Once it is reinforced, it carries over to rules on which the specific action was never tried. How often RL tried the hammer at a given decision barely predicts whether that decision improves; whether its run takes off does.

Rarity sets the odds of this event (Figure 4b). As the switch becomes rarer, fewer runs take off: 21 of 30 at the most frequent level, 10 of 30 at the rarest level where any run takes off, and none below. Which runs take off looks like chance. The same base model may take off with one group size and not with another. And on the same budget, larger groups take off less often: with the most frequent switch, 9 of 10 runs take off with groups of 4 or 16, but only 3 of 10 with groups of 64.

So rarity does not set a wall but a probability, and two runs of the same configuration can end on opposite sides. What decides whether a given run wins the bet? And why do larger groups lose it more often, when they make exactly the same number of attempts? The next section answers both questions with a single count.

\subsection{4. What makes a run take off: counting groups that succeed}\label{what-makes-a-run-take-off-counting-groups-that-succeed}

Recall how the algorithm learns. A group of attempts on one chest pushes the model toward the hammer only if at least one attempt in the group uses the hammer and wins; a group in which every attempt fails teaches nothing (Figure 5a). Before RL begins, the base model tells us how likely this is. For each problem the run will see, we compute the chance that a single attempt switches to the hammer at the right moment and wins, and from it the chance that a group of \emph{N} attempts contains at least one such winner. Adding these chances over every group of the run gives one number: the \emph{expected number of successful groups}.

This number predicts which runs take off. Half of all runs take off when about 35 successful groups are expected; one in ten with about 10, and nine in ten with about 80 (Figure 1). Two simpler counts fail, and seeing why tells us what RL needs.

\begin{figure}[tbp]
\centering
\includegraphics[width=\textwidth]{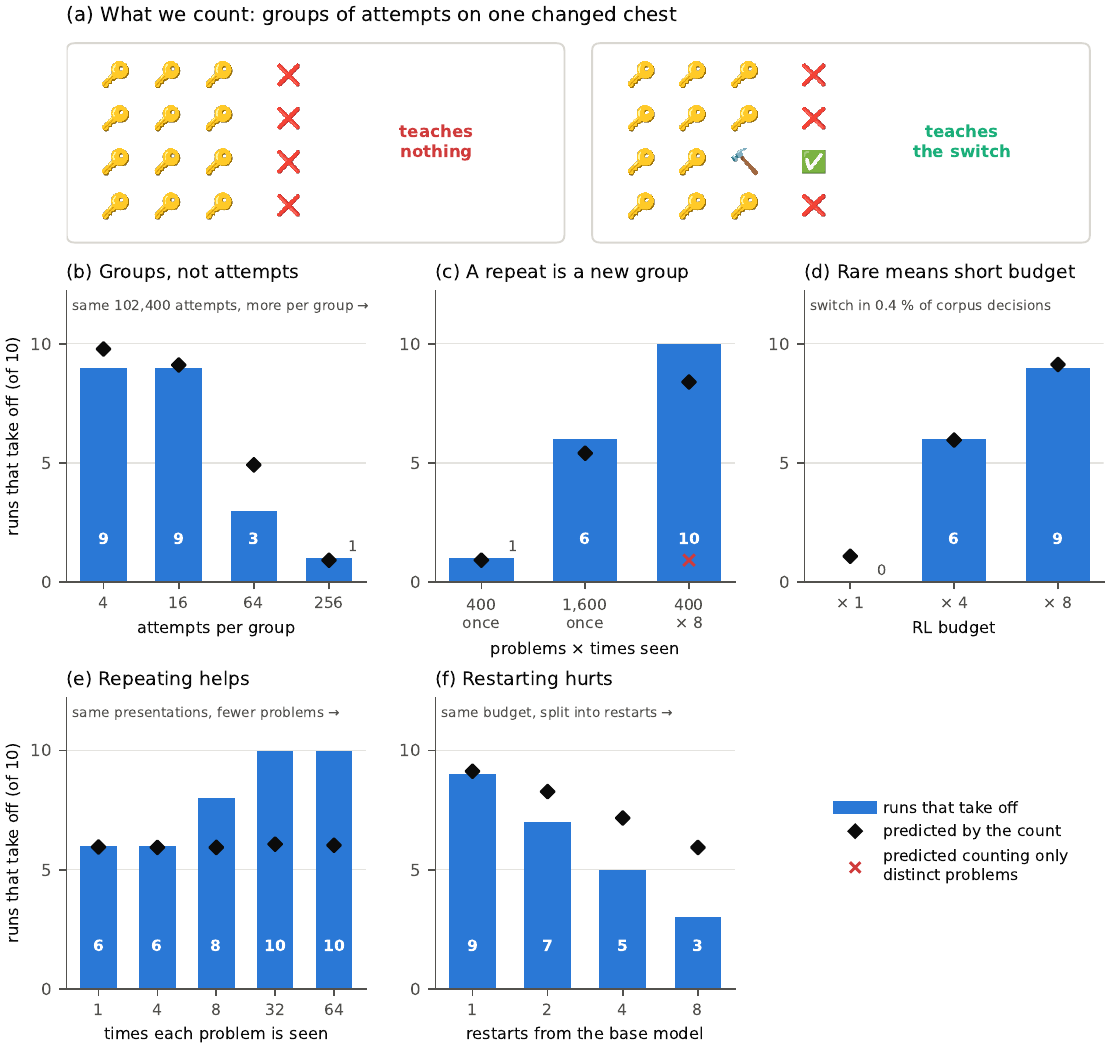}
\caption{\textbf{What makes a run take off.} (a) Two groups of attempts on a changed chest; we count groups like the one on the right. (b)--(f) Runs that take off, of 10 (bars), and the number predicted by the count with the law of Figure 1 (diamonds). (b) The same number of attempts, split into larger groups. (c) The same 400 problems seen once or eight times, and 1,600 problems seen once; the cross is what the law predicts if only distinct problems are counted. (d) A switch that our budget never found, with 4 and 8 times the budget. (e) The same number of groups, spread over fewer problems seen more often. (f) The same budget split into restarts from the base model; bars count seeds in which any restart takes off.}
\label{fig:counting}
\end{figure}

\textbf{Groups, not attempts.} Counting the attempts that discover the hammer cannot explain why larger groups take off less often on the same budget. A group needs only one winner; a second winner in the same group adds little, because the group already points toward the hammer. Groups of 256 attempts on 400 problems make the same 102,400 attempts as groups of 4, and they find the hammer about as often. Yet only 1 of 10 such runs took off, against 9 of 10 with groups of 4 or 16, as the count of groups predicts (Figure 5b).

\textbf{A repeated problem is a new group.} Nor is it enough to count distinct problems. Showing the same 400 problems eight times each, instead of once, raised the take-offs from 1 to 10 of 10 (Figure 5c). Seeing a problem again creates a new group, and so a new chance that some attempt wins with the hammer; counting only distinct problems would predict less than one take-off. The extra training steps are not the reason: spreading 1,600 problems over four times more steps, without repeating them, left the outcome unchanged (6 of 10 in both cases).

\textbf{``Rare'' means ``not enough budget''.} The count also says why the rare switch of Section 3 was never found. With the switch in 1 of 232 corpus decisions, our budget gave too few expected successful groups for any run to take off. With four times the budget, the count predicted about half of the runs to take off, and with eight times most of them. Of 20 runs, 15 took off, against 15.1 predicted (Figure 5d). An action that RL ``could not discover'' was discovered once RL got enough chances to see it succeed. For a switch ten times rarer still, the count asks for about 40 times our budget for half of the runs to take off.

\textbf{Repeating problems helps more than the count says.} With groups of 4 and the same number of groups, a run that sees fewer problems more often takes off more: 6, 6, 8, 10 and 10 of 10 runs with each problem seen 1, 4, 8, 32 and 64 times, where the count predicts about 6 throughout (Figure 5e). The count uses the base model's chance of success on each problem, but that chance grows while the run practices it. After 100 steps, the chance of a winning switch on a problem seen 8 times had grown six-fold, against about 2.5-fold on problems the run had not seen. With repeated problems, the count is a lower bound. Nor does repetition seem to trade generalization for memorization in these runs. Take-off is measured on fixed decision points, not on the training problems. And the chance of a winning switch also grew on 2,000 evaluation problems drawn apart from the training ones: 2.5-fold after 100 steps with each problem seen 8 times, while it fell to 0.7 of its starting value in runs without repeats.

\textbf{Restarting from the base model hurts.} Splitting the same budget into restarts from the base model throws away the progress each run had made toward its first successes. With 1, 2, 4 and 8 restarts, 9, 7, 5 and 3 of 10 seeds took off at least once (Figure 5f). A run that did not take off, on the other hand, leaves little damage: continuing RL from such a run took off in 21 of 30 cases, against 9 of 10 from the base model, a difference we cannot distinguish from chance. The practical advice is to continue the same run rather than to start over.

\textbf{The count is not a quirk of our recipe.} Everything so far used one recipe: one model size, one optimizer, one learning rate and the simplest form of GRPO. We changed it twelve ways, from the learning rate and the optimizer to the form of GRPO (a strong KL penalty, an advantage divided by the group's spread) and the size of the model, from 1.3M to 85M parameters, and repeated seven of the tests above under each change: 910 runs in all. In every setting, counting successful groups predicted take-off better than counting problems or attempts, never by a small margin. What the recipe changes is how many successful groups a run needs. A lower learning rate or plain SGD makes a run need two to three times as many; a KL penalty or a divided advantage, about one and a half times as many; gradient clipping and model size make no difference we can detect (Appendix B).

\subsection{5. Related work}\label{related-work}

\textbf{Knowledge conflicts.} Conflicts between what a model learned and what its context says are well studied in question answering \citep{longpre2021entity,xu2024survey}. Models accept coherent counter-evidence but resist it when the evidence is mixed, more so for popular facts \citep{xie2024chameleon}; they weigh the credibility of sources \citep{schuster2026whose} and can also trust the context too much \citep{huang2025trust}. Measures that compare a model's answer with and without the context \citep{wang2025adacad,bi2026parameters} are close relatives of our switch probability. How often a fact appears in pre-training shapes how well a model knows it \citep{kandpal2023large}. In these studies the prior is estimated rather than set, the evidence is a given text, and nothing is decided over time.

\textbf{Conflicts in agents.} Closest to our work, \citet{xu2026code} change the rules of Baba Is You against familiar physics and find that language models keep acting on the familiar rules. Web agents adopt conflicting memories at the first point where they appear \citep{chen2026compliance}, updating stored facts remains hard \citep{hu2026memoryagentbench}, and code models struggle to use an API after it changes \citep{liu2024codeupdatearena}. None of these studies controls the strength of the prior, knows the right answer under noisy evidence, or compares the same model before and after RL.

\textbf{Post-training and reliance on evidence.} Instruction tuning can reduce reliance on context in conflicts \citep{goyal2025inversion}, and reasoning models override explicit instructions in favour of familiar patterns \citep{jang2025stubborn}. On the other hand, RL fine-tuning makes agents less greedy in bandit tasks \citep{schmied2026greedy}. Our results suggest when each picture applies: RL moves decisions toward whatever is rewarded, in either direction, provided the rewarded action is tried often enough to succeed.

\textbf{Sharpening and discovery.} RL with verifiable rewards improves the first answer but rarely solves problems the base model could not solve with many samples, which suggests that it mostly reweights what the base model already does \citep{yue2025rl}. RL amplifies reasoning behaviours that the base model already shows \citep{gandhi2025cognitive}, and in models pre-trained from scratch on known data it converges on one of the output formats seen in pre-training \citep{zhao2025echo}. In theory, how often the base model samples a good response bounds the cost of exploration in RL fine-tuning \citep{foster2025foundation}, and next-token pre-training shapes exactly this coverage \citep{chen2026coverage}. We set this probability directly through the corpus, over three orders of magnitude, and find that a count of successful groups, not of attempts or problems, tells when RL moves beyond sharpening. Work on how training data decides between learning in the weights and learning in context \citep{chan2022data} suggests that well-exposed rules should resist evidence more; in our setting the opposite held.

\subsection{6. Discussion}\label{discussion}

\textbf{What this means for post-training.} Whether RL ``only sharpens'' what a base model already does is not a property of RL. In our world it depends on one quantity that can be estimated before training: how many groups of attempts are expected to contain a success with the new action. Below the threshold, RL sharpens and the prior wins; above it, RL discovers the action, generalizes it across rules and makes the agent decisive. Near it, the outcome is a matter of luck, so single runs can mislead in both directions. The count also suggests cheap levers when a run does not take off: more budget, problems seen more than once, and continuing the same run rather than restarting it.

\textbf{Why all or nothing?} We have not looked inside the models, but the structure of the world suggests an explanation. The nine test rules and the thirty filler rules share one relational pattern: an obstacle, its default tool, and the default tool of another rule as the alternative. The base model learns the switch from all of them together, so every successful group, on any rule, pushes on the same shared mechanism. Once enough successful groups have accumulated, the change would spread to every obstacle and tool at once. Such sudden, shared transitions resemble the phase change in which transformers form induction heads \citep{olsson2022context} and the abrupt generalization of grokking \citep{power2022grokking,nanda2023progress}. Testing this account with interpretability tools is future work.

\textbf{Limitations.} Our world is synthetic and its language is templated, and the models are small, up to 85M parameters. Prior strength in our models comes from exposure, not from a Bayesian prior. We study one kind of conflict, switching after one's own failures, and one family of RL algorithms, GRPO without a reference model by default; we did not test PPO or other rewards and horizons. The threshold of about 35 successful groups is specific to this world and recipe. Our single fit across the thirteen recipes predicts a left-out recipe that lies between fitted ones (higher learning rates, no clipping, the divided advantage) but not one beyond them (lower learning rates, new model sizes), so it describes the settings we ran rather than predicting new ones.

\textbf{Future work: a real pre-trained model.} We took a first look at OLMo-3-7B \citep{olmo2025olmo3}, where the prior comes from real pre-training, using four public checkpoints of the same pipeline (Appendix C). On single decisions, the same measure behaves as expected: the switch after one failure is smaller for rules with a stronger prior, and post-training makes the model switch more after a failure. But the model does not yet play whole games reliably, so we could not run RL on it or count its successful groups. Doing so, in a model large enough to play, and in worlds other than ours, is the natural next test of the count.

\subsection{7. Conclusion}\label{conclusion}

We asked who wins when an agent's prior and its evidence disagree, and how outcome-reward RL changes the answer. In a world where both sides are known, base models hedge as their training data did, and RL turns that hedging into decisions, siding with the evidence when the evidence should win. It can do so only for actions it sees succeed: a run learns a new action once it has seen enough groups of attempts in which that action won, a count we can compute before training and that held across learning rates, optimizers, GRPO variants and model sizes. RL can only reinforce what it sees succeed; how often it sees it is something we can measure, and change.

\bibliographystyle{plainnat}
\bibliography{references}

\clearpage

\clearpage

\subsection{Appendix A. Methods in detail}\label{appendix-a.-methods-in-detail}

\subsubsection{A.1 The game}\label{a.1-the-game}

The puzzle engine treats rules as data. A game takes place in one room whose exit is blocked by one to three obstacles. Each obstacle has exactly two candidate tools next to it, named in random order. The agent can \texttt{take} a tool, \texttt{use} a tool on an obstacle, and \texttt{go\ exit} once no obstacle remains, and the engine lists the valid actions at every step. There are 9 test rules and 30 filler rules, each pairing an obstacle \emph{O} with its default tool \emph{D}, such as the locked chest and the key.

In the normal world, using \emph{D} on \emph{O} removes it with probability \(1 - \varepsilon\), with \(\varepsilon = 0.2\), and otherwise does nothing; the other candidate \emph{A} does nothing. In a changed world, \emph{D} never works and \emph{A} removes the obstacle. Every failure produces the same observation, ``nothing''. Each obstacle follows its changed rule with probability \(\nu\), which gives the posterior \(\pi(k)\) of Section 2. When the next attempt is the last one, \emph{D} wins with probability \((1 - \pi)(1 - \varepsilon)\) and \emph{A} with probability \(\pi\), so switching is better once \(\pi(k) > \tau = (1 - \varepsilon)/(2 - \varepsilon) \approx 0.44\). In each game, \emph{A} is drawn from the default tools of rules that are not in the game, so that both candidates are ordinary tools and only the learned link between \emph{O} and \emph{D} tells them apart.

\subsubsection{A.2 Decision points, the switch probability and take-off}\label{a.2-decision-points-the-switch-probability-and-take-off}

We measure models at decision points with a fixed prefix. For a test rule, we build a changed-world game and write a prefix in which the agent takes both tools, in random order, and then uses \emph{D} on \emph{O} \emph{k} times, each time with ``nothing'' as the outcome; \emph{k} runs from 0 to 3. The game also holds zero to two filler obstacles, already solved in the prefix or still pending. We draw 100 such instances per rule and \emph{k}, which differ in \emph{A}, the filler obstacles and the order of items.

Every measurement in the small models is exact. We score each valid action by teacher forcing, summing the log-probabilities of its tokens up to and including the separator that ends an action, and compute \(s(k)\) from the probabilities of ``use \emph{A} on \emph{O}'' and ``use \emph{D} on \emph{O}'', averaged over instances. Base models and their RL descendants are measured on the same instances, so every change is paired. We report the distance to the ideal belief, \(|s - \pi|\), and the distance to the best decision, \(|s - \mathbf{1}[\pi > \tau]|\), separately.

A rule is \emph{learned} in an RL run if its switch probability after two or after three failures rises by more than 0.05 from the base model, with a paired 95 \% confidence interval over the 100 instances. A run \emph{takes off} if it learns more than half of the nine test rules. We use this definition for every run in the paper.

\subsubsection{A.3 Corpora and base models}\label{a.3-corpora-and-base-models}

Games are written in a templated language with one token per word, 108 tokens in all, for example \texttt{\textless{}game\textgreater{}\ room\ locked\_chest\ :\ key\ hammer\ \textless{}/room\textgreater{}\ \textgreater{}\ take\ key\ \textbar{}\ ok\ \textgreater{}\ use\ key\ locked\_chest\ \textbar{}\ nothing\ \textgreater{}\ ...}. Each corpus has 9 million games, between 516 and 548 million tokens, generated once and seen once. The share \emph{f} of games that contain a test rule sets its exposure: 90,000, 9,000 or 900 games for \emph{f} = \(10^{-2}\), \(10^{-3}\) and \(10^{-4}\). Filler rules change with the same probability \(\nu\), so the general skill of switching is learned from about 900,000 changed-world examples at \(\nu = 0.05\), not from the test rules alone. In the rarity corpora the late player plays the filler rules too, so only a share \(\rho\) of these examples shows an early switch.

The \textbf{Bayesian player} knows \(\nu\) and \(\varepsilon\) but not the world. At each obstacle it uses \emph{D} first and, after \emph{k} failures, switches to \emph{A} with probability \(\pi(k)\), returning to \emph{D} if \emph{A} fails. A model that learned this corpus perfectly would have \(s(k) = \pi(k)\) for every rule. In the rarity corpora, a share \(1 - \rho\) of the games is played instead by a \textbf{late player} that switches only from the fourth failure on, beyond every decision that counts in RL; the corpora of one seed share their games and differ only in the player. After two failures the corpus then shows the switch in a share of about \(0.44\,\rho\) of decisions, from 1 in 23 at \(\rho\) = 0.1 to 1 in 23,000 at \(\rho\) = \(10^{-4}\). The seven levels are \(\rho\) = 0.1, 0.05, 0.03, 0.02, 0.01, \(10^{-3}\) and \(10^{-4}\).

The models follow nanoGPT \citep{karpathy2022nanogpt}: 6 layers, 6 heads, width 384, context 256, tied embeddings, no biases, 10.7M parameters. The two other sizes have 3 layers, 3 heads and width 192 (1.3M parameters), and 12 layers, 12 heads and width 768 (85M). Games are packed into chunks of whole games, so that every game is seen exactly once. Training uses AdamW \citep{loshchilov2019decoupled} with \(\beta = (0.9, 0.95)\) and weight decay 0.1 on matrices, a learning rate of \(10^{-3}\) with 500 warm-up steps and cosine decay to \(10^{-4}\), batches of 256 chunks, about 9,000 steps and gradient clipping at 1.0, in bfloat16 on NVIDIA A6000 GPUs. Each seed has its own corpus. Every base model passed two controls: in 200 sampled normal-world games with one to three obstacles it left the room at least 96.5 \% of the time, with every loss an invalid action rather than a timeout; and its validation loss changed by less than 0.3 \% over the last fifth of training. The two other model sizes passed the same controls. A further control showed that estimating \(s(k)\) from 20 sampled episodes per decision is unbiased against the exact measure.

\subsubsection{A.4 Outcome-reward RL}\label{a.4-outcome-reward-rl}

We post-train with GRPO \citep{shao2024deepseekmath}. Each problem is one obstacle of a test rule, chosen uniformly, in a changed world with probability \(\nu\) and with noise \(\varepsilon\), the same distribution as in pre-training. The model plays \emph{N} attempts at temperature 1; the advantage of an attempt is its reward minus the mean reward of its group, without dividing by the standard deviation \citep{liu2025understanding}; and one gradient step follows each batch, so the policy ratio is 1 and nothing is clipped. There is no KL penalty and no reference model. The optimizer is AdamW with learning rate \(3 \times 10^{-5}\), \(\beta_2 = 0.95\), no weight decay and gradient clipping at 1.0. An unreadable or invalid action ends the episode with reward 0.

The reward is 1 if the agent leaves the room within a step limit drawn uniformly from 4 to 7 for each problem, and 0 otherwise. With a limit of \emph{k} + 4, the decision after \emph{k} failures is the last chance, where the reward-maximizing choice is exactly to switch when \(\pi(k) > \tau\). The corpus games allow 20 steps, which would let every attempt win and leave every group without signal. Unless stated otherwise, a run takes 200 steps of 512 attempts, 102,400 attempts in all, so that larger groups see fewer problems. We evaluate the reward on 2,000 held-out problems against the optimal policy for the same horizon, computed by dynamic programming. During RL we log, for each rule and number of failures, how often the decision is reached and how often the model tries \emph{A} there.

\subsubsection{A.5 Counting successful groups}\label{a.5-counting-successful-groups}

For each base model we measure its \emph{reach}: on a problem in a changed world, the probability \emph{q} that one attempt, played at temperature 1 as in RL, switches to \emph{A} after two or three failures and then wins. Because \emph{A} can be as rare as \(10^{-5}\) per decision, sampling would almost never see it. The estimator scores exactly, at every decision point, the probability of using \emph{A} and then leaving, adds it to \emph{q} weighted by the probability of not having switched earlier, and samples only the rest of the attempt with \emph{A} excluded. It is unbiased, and our tests check it against direct sampling. We measure \emph{q} on 2,000 problems common to all models and on the problems each run will actually see.

From the reach on each of a run's problems we compute three counts, all before training. Let \(q_i\) be the reach on the \emph{i}-th distinct problem of the run, \emph{N} the group size and \emph{r} the number of times each problem is presented. Counting \textbf{attempts} gives the expected number of winning switches; counting \textbf{problems}, the expected number of distinct problems with at least one winning attempt; counting \textbf{groups}, the expected number of presentations whose group contains at least one winning attempt:

\[D_\text{att} = \sum_i r N q_i, \qquad D_\text{prob} = \sum_i \bigl(1 - (1 - q_i)^{rN}\bigr), \qquad D_\text{rep} = \sum_i r \bigl(1 - (1 - q_i)^{N}\bigr).\]

When no problem repeats (\emph{r} = 1), \(D_\text{rep}\) equals \(D_\text{prob}\). The take-off law is

\[P(\text{take-off}) = 1 - \exp\bigl(-\exp(b_0 + b_1 \log_{10} D)\bigr),\]

fitted by maximum likelihood. Fitted to the 400 runs that start from a base model, with \(D = D_\text{rep}\), it gives \(b_0 = -5.75\) and \(b_1 = 3.47\), and even odds at about 35 successful groups; this is the line of Figure 1 and the source of every prediction in Figure 5. We compare counts by the log-likelihood of the take-offs they predict.

\subsubsection{A.6 The runs}\label{a.6-the-runs}

\begin{table}[htbp]
\centering
\small
\renewcommand{\arraystretch}{1.15}
\begin{tabular}{@{}p{3.4cm}p{6.0cm}p{2.3cm}p{1.4cm}r@{}}
\toprule
Set of runs & What changes & $\rho$ & Group size $N$ & Runs \\
\midrule
First experiment & 3 exposures; $\nu = 0.05$ or $0.3$; 2 objectives $\times$ 3 group sizes & 1 & 4, 16, 64 & 120 \\
Exposure $\times$ reliability & 3 exposures $\times$ 3 reliabilities in each base model (Figure 3) & 1 & 16 & 18 \\
Rarity & 7 levels of rarity $\times$ 3 group sizes, 200 steps (Section 3) & $10^{-4}$ to $10^{-1}$ & 4, 16, 64 & 210 \\
More budget & 4$\times$ budget, as more problems per step or as more steps; longer AdamW memory & $10^{-1}$ & 64 & 30 \\
Groups of 256 & 1$\times$, 2$\times$ and 4$\times$ budget & $10^{-1}$, $5 \times 10^{-2}$ & 256 & 40 \\
Repeats and budget & 400 problems $\times$ 8; 1,600 problems over 4$\times$ more steps; 4$\times$ and 8$\times$ budget & $10^{-1}$, $10^{-2}$ & 4, 256 & 40 \\
Small pools & problems repeated 4 or 8 times & $2 \times 10^{-2}$ to $10^{-1}$ & 64, 256 & 40 \\
Long runs & 1,400 steps from runs that did not take off or from the base model; 4 restarts of 400 steps & $10^{-2}$ & 4 & 80 \\
Restarts and repeats & 2 restarts of 800 or 8 of 200 steps; each problem seen 4 to 64 times & $10^{-2}$ & 4 & 140 \\
Twelve recipes & 7 arms from the rows above under each change of recipe (Appendix B) & $10^{-2}$, $10^{-1}$ & 4, 256 & 840 \\
\midrule
Total & & & & 1,558 \\
\bottomrule
\end{tabular}
\caption{\textbf{Every RL run in this paper.} $\rho = 1$: every corpus game is played by the Bayesian player. From the rarity runs on, each arm has 10 runs, one per base model.}
\label{tab:runs}
\end{table}

From the rarity runs on, the RL seed equals the seed of the base model. The first experiment varied exposure only, with nine test rules at three exposure levels and ten base models per reliability; the six combinations of objective (the binary reward and a discounted one) and group size, and the gap closed on held-out problems in Section 3, come from it. All training and RL runs used NVIDIA A6000 GPUs in bfloat16.

\subsection{Appendix B. The count across thirteen settings}\label{appendix-b.-the-count-across-thirteen-settings}

We changed the recipe twelve ways: the learning rate two- and four-fold lower and higher, no gradient clipping, plain SGD instead of AdamW, the advantage divided by the group's standard deviation, a KL penalty toward the base model with \(\beta = 1\) \citep{ouyang2022training}, the last two combined, and models of 1.3M and 85M parameters. A penalty of \(\beta = 1\) is far above the values common in RL with verifiable rewards, such as 0.04 \citep{shao2024deepseekmath}, so it is a strong stress test that pulls the model back toward its prior. Each setting repeats seven tests of Section 4, 70 runs per setting and 910 with our own recipe. In each setting, counting successful groups beat counting problems and attempts, by at least 12 points of log-likelihood. The curve kept its slope in ten settings; it is flatter under SGD and steeper under the KL penalty.

\begin{figure}[htbp]
\centering
\includegraphics[width=0.7\textwidth]{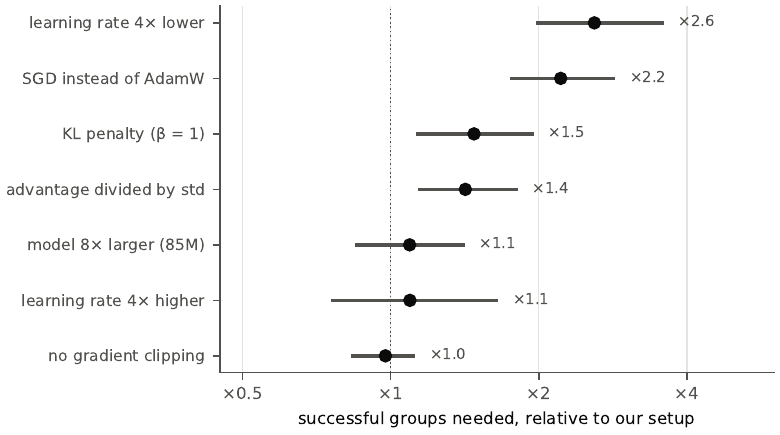}
\caption{\textbf{What each change of recipe costs.} How many more successful groups a run needs, relative to our recipe, for each change within the range we tested; 95\,\% bootstrap intervals over the 20 pre-training corpora. A KL penalty and a divided advantage, applied together, cost about as much as neither.}
\label{fig:law}
\end{figure}

To compare the settings, we fitted one curve to all 910 runs, with a single slope and a shift of the threshold for each change. Giving each setting its own slope does not improve it (p = 0.46). Figure \ref{fig:law} converts each shift into the change we actually ran. When one setting is left out and predicted from the others, the prediction works for settings between fitted ones (higher learning rates, no clipping, the divided advantage) and fails beyond them (lower learning rates, new model sizes), so the fit describes these settings but does not yet predict new ones.

\subsection{Appendix C. OLMo-3-7B}\label{appendix-c.-olmo-3-7b}

We repeated the decision-point measure in OLMo-3-7B \citep{olmo2025olmo3}, where the prior comes from real pre-training, with four public checkpoints of one pipeline: the base model, SFT, DPO and the final Instruct model. A renderer turns each game into a chat in natural language; a failed use returns only ``Nothing happens.'', and the system message states that even the right tool fails about 20 \% of the time. Each action is scored exactly as the model's next reply.

The stronger a rule's prior, the less the final model switches after one failure. Across the nine test rules, the rank correlation between prior strength and the switch after one failure is −0.97, and the model switches after one failure 95 \% of the time for weak rules and 74 \% for strong ones. Post-training spreads the rules apart, from 0.3 orders of magnitude of prior strength in the base model to 3.8 in the final model, and makes the model switch more after one failure. Part of this effect is the starting point: stronger rules are also less likely to be switched before any failure.

No checkpoint plays whole games well enough for RL: the final model wins 60 \% of normal-world games, and almost every loss is a reply that is not a valid action. OLMo-3-7B therefore answers single decisions as our account predicts, but it does not yet play well enough for RL on our games, which is why it appears in this paper only as future work.

\end{document}